\documentclass[11pt]{article}
\usepackage{coling2020}
\usepackage{times}
\usepackage{url}
\usepackage{latexsym}
\usepackage{CJKutf8}
\usepackage{graphicx}
\usepackage{array}
\usepackage{multirow}
\usepackage{amsfonts}
\usepackage{amstext}
\usepackage{amsmath}
\usepackage{floatrow} 
\usepackage{subfigure}
\newfloatcommand{capbtabbox}{table}[][\FBwidth]

\usepackage{microtype}

\colingfinalcopy 

\title{Porous Lattice Transformer Encoder for Chinese NER}

\author{
 Mengge Xue\textsuperscript{1,2}, Bowen Yu\textsuperscript{1,2}, Tingwen Liu\textsuperscript{1,2}\thanks{{} {} Corresponding author.}, Yue Zhang\textsuperscript{3} \\
 \bf Erli Meng\textsuperscript{4} \and Bin Wang\textsuperscript{4}\\
 \textsuperscript{1}Institute of Information Engineering, Chinese Academy of Sciences, China \\
 \textsuperscript{2}School of Cyber Security, University of Chinese Academy of Sciences, China \\
 \textsuperscript{3}School of Engineering, Westlake University, Hangzhou, China \\
 \textsuperscript{4}Xiaomi AI Lab, Xiaomi Inc., Beijing, China \\
 {\tt \{xuemengge, yubowen, liutingwen\}@iie.ac.cn}\\
{\tt yue.zhang@wias.org.cn}\\
{\tt \{wangbin11, mengerli\}@xiaomi.com}
}
\date{}

\begin{document}
\maketitle
\begin{abstract}
Incorporating lexicons into character-level Chinese NER by lattices is proven effective to exploit rich word boundary information.
Previous work has extended RNNs to consume lattice inputs and achieved great success.
However, due to the DAG structure and the inherently unidirectional sequential nature, this method precludes batched computation and sufficient semantic interaction.
In this paper, we propose PLTE, an extension of transformer encoder that is tailored for Chinese NER, which models all the characters and matched lexical words in parallel with batch processing.
PLTE augments self-attention with positional relation representations to incorporate lattice structure.
It also introduces a porous mechanism to augment localness modeling and maintain the strength of capturing the rich long-term dependencies.
Experimental results show that PLTE performs up to 11.4 times faster than state-of-the-art methods while realizing better performance.
We also demonstrate that using BERT representations further substantially boosts the performance and brings out the best in PLTE. 
\end{abstract}
\section{Introduction}

Named Entity Recognition (NER) is a fundamental task in natural language processing (NLP), which aims to automatically discover named entities and identify their corresponding categories from plain text.
NLP tasks  such as information retrieval~\cite{Berger:2017SIGIR}, relation extraction~\cite{Yu:2019IJCAI} and  entity linking~\cite{Xue:2019IJCAI} require the NER as one of their preprocessing components. Recent studies show that English NER models have achieved improved performance by integrating character information into word representations based on sequence labeling.
Different from English NER, East Asian languages including Chinese are written without explicit word boundary.
One intuitive way to solve this problem is to segment the input sentences into words first, and then to apply word sequence labeling \cite{Yang:2016CICLing,He:2017AAAI}.
However, such methods suffer from error propagation between these two subtasks.

\begin{figure*}[t]
\centering
\includegraphics[width=0.9\textwidth]{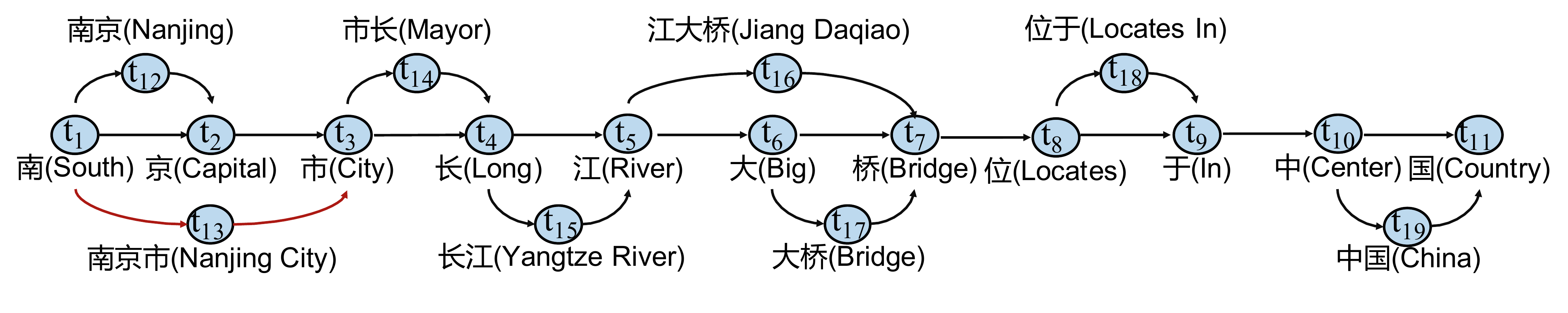}
\caption{
\label{difference-between-attention}
Example of word character lattice. 
\begin{CJK}{UTF8}{gbsn}
Restricted by the unidirectional sequential nature, lattice LSTM cannot model the semantic interaction between word ''南京市(Nanjing City)" and its constituent characters ''南(South)" and ''京(Capital)", resulting in the loss of crucial information for tagging.
Besides, lattice LSTM cannot perform batched computation due to the directed acyclic graph input structure.
\end{CJK}}
\end{figure*}

To overcome this limitation, efforts have been devoted to incorporating word information by leveraging lexicon features~\cite{Peng:EMNLP2015,Cao:2018AAAI,Wu:2019WWW}. As recent state-of-the-art (SOTA) lattice-based method, \newcite{Zhang:ACL2018} integrated matched lexical words information into character sequence with a directed acyclic graph (DAG) structure using lattice LSTM.
While obtaining promising results, this model faces two challenges.
\textbf{First}, as a extension to the non-parallelizable sequential LSTM to a DAG structured model, lattice LSTM is restricted to preprocess one character at a time, which can make it infeasibly to deploy.
\textbf{Second}, due to the inherently unidirectional sequential nature, lattice LSTM fails to incorporate the word-level semantics  into the representation of the characters except for the last character in each word, despite that such information can be crucial for character-level sequence tagging.
Taking the sentence in Figure~\ref{difference-between-attention} as an example, lattice LSTM decodes the information of the lexical word 
\begin{CJK}{UTF8}{gbsn}
``南京市(NanJing City)" to ``市(City)" but skips the other two inside characters ``南(South)" and ``京(Capital)", although the semantics and boundary information of ``南京市(NanJing City)" can be useful knowledge for predicting the tag of ``南(South)"
\end{CJK}
as ``B-LOC".

In this paper, we address these issues by considering a novel \textbf{P}orous \textbf{L}attice \textbf{T}ransformer \textbf{E}ncoder (PLTE).
Inspired by previous research on machine  translation~\cite{Xiao:ACL2019,Sperber:ACL2019}, which integrated lattice-structured inputs into self-attention models, we propose a lattice transformer encoder for Chinese NER by introducing lattice-aware self-attention, which borrows the idea from the relative positional embedding~\cite{shaw:2018naacl} to make self-attention aware of the relative position information in lattice structure.
Considering that self-attention network calculates attention weights between each pair of tokens in a sequence regardless of their distance, we simply concatenate all the characters and lexical words as input to consume lattices without resorting to the DAG structure.
In this way, characters coupled with lexical words can be processed in batches.
A lexical word representation is allowed to build a direct relation with the included characters by lattice-aware self-attention, thus addressing the second issue. 

Some work~\cite{Yang:2018EMNLP,Yang:2019NAACL} demonstrates that self-attention benefits from locality modeling,
especially for the NER task. 
As we can see from the example in Figure~\ref{difference-between-attention}, 
\begin{CJK}{UTF8}{gbsn}
the word ``位于(Locates In)" is the immediate and most obvious feature to guide the neighboring character ``桥(Bridge)"
\end{CJK}
to be identified as ``E-LOC" instead of ``E-PER", while 
\begin{CJK}{UTF8}{gbsn}
"中国(China)" 
\end{CJK}
has no contribution to this decision. Given this observation, we further introduce a novel porous mechanism to enhance the local dependencies among neighboring tokens.
The key insight is to modify the self-attention architecture by replacing the fully-connected topology with a pivot-shared structure.
In this particular, every two non-neighboring tokens are connected by a shared pivot node to strengthen the dependency for two neighboring tokens. Experimental results on four datasets demonstrate that our model performs up to 11.4 times faster than baselines and achieves better performance. Furthermore, we show that our model can be easily integrated into the pre-trained language model such as BERT~\cite{devlin:2019naacl}, and combining them further improves the state of the art.

In summary, this paper makes the following contributions: 
(1) We investigate lattice transformer encoder for Chinese NER, which is capable of handling lattices in batch mode and capturing dependencies between characters and matched lexical words. 
(2) We revise lattice-aware attention distribution via a porous mechanism, which enhances the ability of capturing useful local context.
(3) Experimental results show that the proposed model is effective and efficient. The source code of this paper can be obtained from https://github.com/strawberryx/PLTE.
\section{Related Work}

Our work is in line with NER models based on neural networks and lattice transformer models.

\newcite{Huang:2015arXiv} proposed a BiLSTM-CRF model for NER and achieved strong performance.
\newcite{Santos:2015NEWS} used word- and character-level representations based on the CharWNN deep neural network.
\newcite{Lample:2016NAACL} designed a character LSTM and word LSTM for NER. 
Compared to our work, these word-based methods suffer from segmentation errors.

To avoid segmentation errors, most recent NER models are built upon character sequence labeling.
\newcite{Peng:EMNLP2015} proposesd a joint training objective for three types of neural embeddings to better recognize entity boundary.
\newcite{Lu:LREC2016} presented a position-sensitive skip-gram model to learn multi-prototype Chinese character embeddings.
\newcite{He:2017AAAI} took the positional character embeddings into account.
Although these methods achieve promising performance, they ignore word information lying in character sequence.

Some work exploits rich word boundary and semantic information in character sequence.
\newcite{Cao:2018AAAI} applied an adversarial transfer learning framework to integrate the task-shared word boundary information into Chinese NER.
\newcite{Liu:NAACL2019} explored four different strategies for Word-Character LSTM.
\newcite{Gui:2019IJCAI} proposed a CNN-based NER model that incorporates lexicons using a rethinking mechanism.
Recent state-of-the-art methods exploit lattice-structured models to integrate latent word information into character sequence, which has been proven effective on various NLP tasks~\cite{su:2017aaai,tan:2018nc} .
Specifically,
\newcite{Zhang:ACL2018} utilized the lattice LSTM to leverage explicit word information over character sequence labeling. Based on this method,
\newcite{gui:2019emnlp} and \newcite{sui:2019emnlp} formulated the lattice structure as a graph and leveraged Graph Neural Networks (GNNs) to integrate lexical knowledge. However, for the NER task, coupling pre-trained language models such as BERT~\cite{devlin:2019naacl} with GNNs and fine-tuning them can be non-trivial.

Lattice transformer has been exploited in NMT~\cite{Xiao:ACL2019}, as well as speech translation~\cite{Sperber:ACL2019,Zhang:ACL2019}. Compared with existing work, our proposed porous lattice transformer encoder is different in both motivation and structure.
We revise the fully-connected attention distribution with a pivot-shared structure via the porous mechanism to enhance the local dependencies among neighboring tokens.\footnote{Differences between lattice self-attention and porous lattice self-attention are shown in Figure 1 in the Appendix.}
To our knowledge, we are the first to design a lattice transformer for Chinese NER.

\section{Background}
In this section, we first briefly review the self-attention mechanism, then move on to current lattice Transformer that our PLTE model is built upon.

\subsection{Self-Attention}
Self-attention mechanism has attracted increasing attention due to their flexibility in parallel computation and dependency modeling. 
Given an input sequence representation $\boldsymbol{\rm{X}} =\{\boldsymbol{\rm{x }}_1, \cdots,\boldsymbol{\rm{x }}_n\} \in \mathbb{R}^{n\times d}$, we can first transform it into queries $\boldsymbol{\rm{Q}}=\boldsymbol{\rm{X}}\boldsymbol{\rm{W}}^Q\in \mathbb{R}^{n\times d_k}$, keys $\boldsymbol{\rm{K}}=\boldsymbol{\rm{X}}\boldsymbol{\rm{W}}^K\in \mathbb{R}^{n\times d_k}$, and values $\boldsymbol{\rm{V}}=\boldsymbol{\rm{X}}\boldsymbol{\rm{W}}^V\in \mathbb{R}^{n\times d_v}$, where $\{\boldsymbol{\rm{W}}^Q, \boldsymbol{\rm{W}}^K, \boldsymbol{\rm{W}}^V\}$ are trainable parameters. The output sequence representation is calculated as:
\begin{equation}
\small
{\rm{Att}}(\boldsymbol{\rm{Q}},\boldsymbol{\rm{K}},\boldsymbol{\rm{V}}) = {\rm{Softmax}}(\frac{\boldsymbol{\rm{Q}}\boldsymbol{\rm{K}}^T}{\sqrt{d_k}})\boldsymbol{\rm{V}},
\end{equation}
where $\sqrt{d_k}$ is the scaling factor. 

\begin{figure*}[htbp]
\subfigure[]{
\label{Ourmodel_Framework_Fig}
\includegraphics[width=0.48\textwidth]{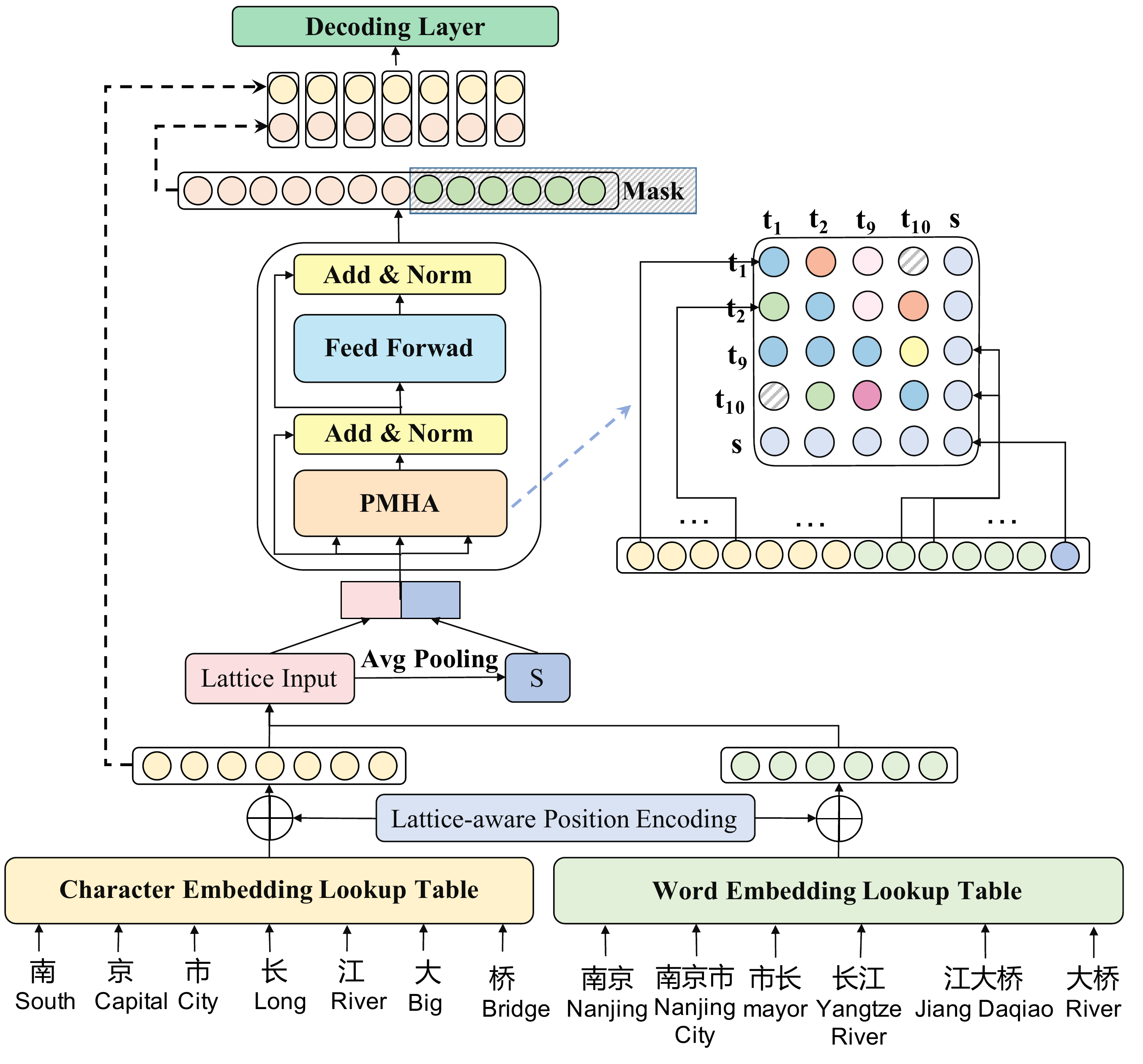}
}
\subfigure[]{
\label{relation-example}

\includegraphics[width=0.48\textwidth]{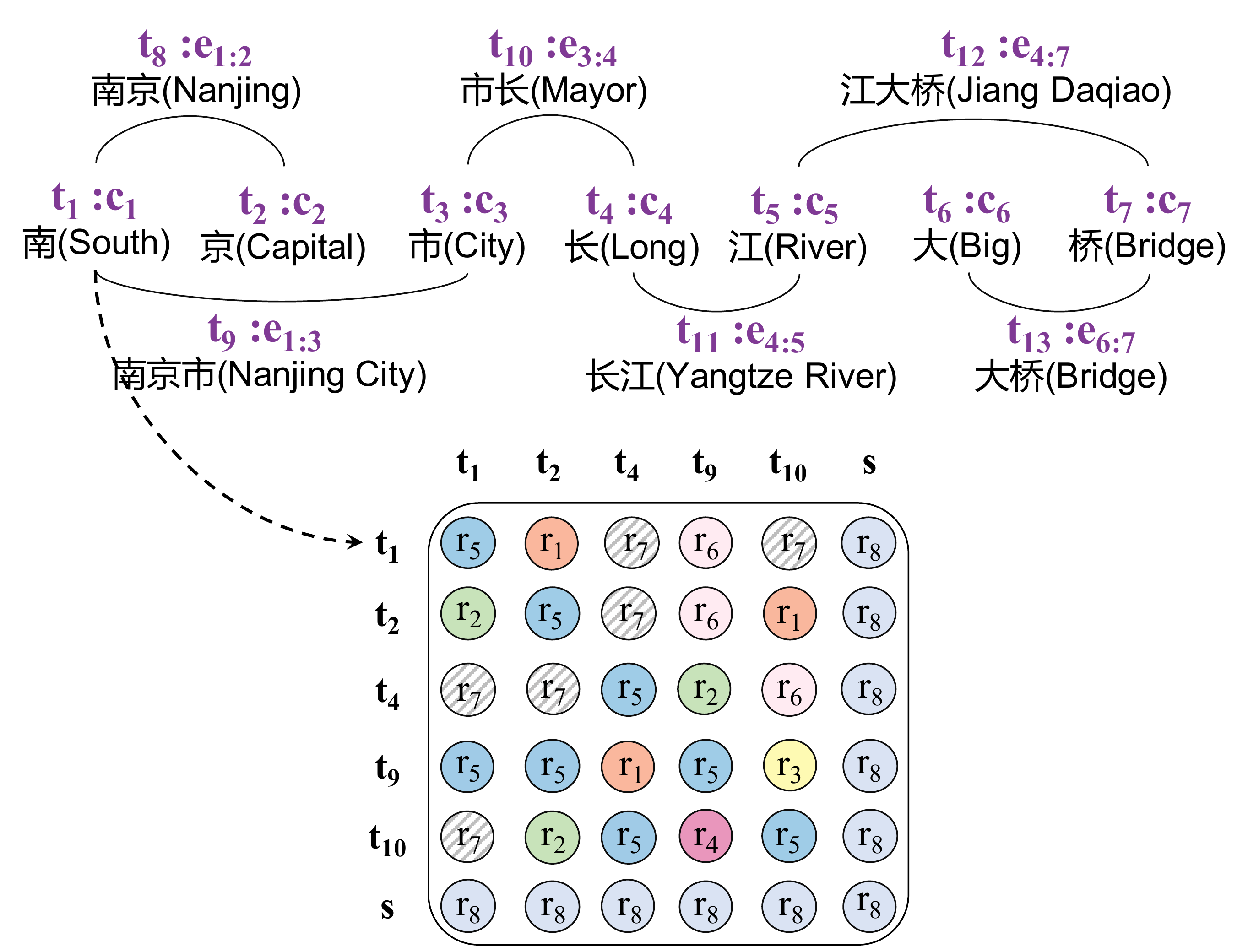}
}%
\centering
\caption{
(a) The overall architecture of PLTE (best viewed in color). Characters and lexical words are shown in yellow and green, respectively.
We concatenate character and word embeddings as lattice input. 
When decoding, we mask words and just make sequence labeling for characters;
and (b) Illustration of the relative position relation matrix. 
Notice that we present several relations among partial tokens as instances. 
Different colors indicate different relations defined in Figure \ref{relation}. 
For instance, the relation between $t_4$ and $t_{10}$ is $r_6$, since that 
\begin{CJK}{UTF8}{gbsn}
``长(Long)" is included in ``市长(Mayor)". The circle filled with lines denotes that we don't compute attention between non-neighboring tokens due to our porous mechanism.
\end{CJK}
}
\end{figure*}

\subsection{Lattice Transformer}
Transformer has been used for many NLP tasks, notably machine translation and language modeling~\cite{wang:2019acl,devlin:2019naacl}. By invoking multi-layer self-attention for global context modeling, Transformer enables paralleled computation and addresses the inherent sequential computation shortcoming of RNNs.
Lattice Transformer is a generalization of the standard transformer architecture to accept lattice-structured inputs, it linearizes the lattice structure and introduces a position relation score matrix to make self-attention aware of the topological structure of lattice:
\begin{equation}
\small
\label{equ:self-att}
{\rm{Att}}(\boldsymbol{\rm{Q}},\boldsymbol{\rm{K}},\boldsymbol{\rm{V}}) = {\rm{Softmax}}(\frac{\boldsymbol{\rm{Q}}\boldsymbol{\rm{K}}^T+\boldsymbol{\rm{R}}}{\sqrt{d_k}})\boldsymbol{\rm{V}},
\end{equation}
where $\boldsymbol{\rm{R}} \in \mathbb{R}^{n\times n}$ encodes the lattice-dependent relations between each pair of elements from the lattices, and its computational method relies on the specific relation definition according to the task objective.

\section{Models}
The overall structure of our model is shown in Figure~\ref{Ourmodel_Framework_Fig}, which consists of 3 main components, lattice input layer, Porous lattice transformer encoder and BiGRU-CRF decoding.
\subsection{Lattice Input Layer}
The input layer aims to embed both semantic information and position information of tokens into their token embeddings.

\paragraph{Word-Character Embedding}
Formally, let $S = \{c_1,...,c_M\}$ denotes a sentence, where $c_i$ is the $i$-th character.
The lexical words in the lexicon that match a character subsequence can be formulated as $e_{i:j}$, where the index of the first and last letters are $i$ and $j$, respectively.
Similarly, we can also represent $c_i$ as  $e_{i:i}$. As shown in the top half of Figure \ref{relation-example}, $e_{3:4}$ indicates the lexical word 
\begin{CJK}{UTF8}{gbsn}
named ``市长(Mayor)" which contains $c_3$ named ``市(City)" and $c_4$ named ``长(Long)". Each character $c_i$ can be turned into the vector $\boldsymbol{\rm{x}}_i^c$ which includes it's character embedding and bigram embedding. By looking up the vector from a pre-trained word embedding matrix, each matched lexical word $e_{i:j}$ is represented as a vector $\boldsymbol{\rm{x}}_{i:j}^w$.
\end{CJK}

\paragraph{Lattice-Aware Position Encoding}
Since self-attention architecture contains no recurrence, to make the model aware of the sequence order, we add position embedding to the semantic embedding of each token.
Specifically, the position of a character is defined as its absolute position in the input sequence $S$.
And the position of a matched word is the position of its first character.
For example, in Figure \ref{relation-example}, the position of word
\begin{CJK}{UTF8}{gbsn}
``南京(Nanjing)" is $1$ because this sentence begins with ``南(South)".
\end{CJK}

Finally, since position information is incorporated into token embeddings, we can simply put the matched words to the 
end of the character sequence $S$ and form a new token sequence $T=\{t_i\}_{i=1}^N$ to consume lattice structure, where $N$ is the sum of the number of characters and words. 
See the top half of Figure \ref{relation-example} for the detailed correspondence.

\subsection{Porous Lattice Transformer Encoder}

As mentioned in the Introduction, our primary goal is to adapt the standard transformer to the task of Chinese NER with lattice inputs.
To this end, we first propose lattice-aware self-attention to consume input tokens and the relative position information of lattice structure. 
\begin{figure}[t]
\centering
\includegraphics[width=0.9\textwidth]{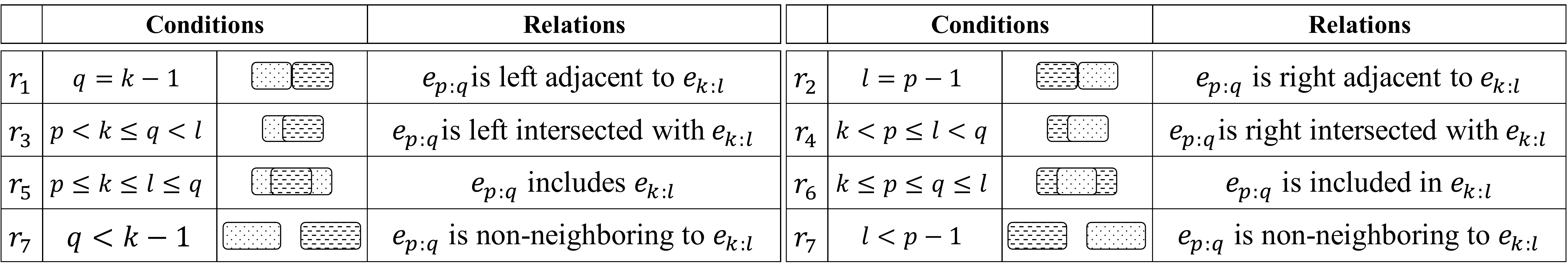}
\caption{
\label{relation}
Relation between $e_{p:q}$ and $e_{k:l}$. 
We use the block filled with dots and lines to present $e_{p:q}$ and $e_{k:l}$, respectively. Notice that if $p\!=\!q\!=\!k\!=\!l$, we denote the relation between $e_{p:q}$ and $e_{k:l}$ as $r_5$. And relation $r_7$ consists of two cases.
}
\end{figure}
Then, we design a porous mechanism which learns sparse attention coefficients by replacing the fully-connected topology with a pivot-shared structure to enhance the association between neighboring elements.
We also use multi-head attention~\cite{vaswani:NIPS2017} to capture information from different representation subspaces jointly.

\paragraph{Lattice-Aware Self-Attention (LASA)}
The position embedding method described above only indicates the sequential order and cannot capture the relative position information of the lattice-structured input. 
For example, in Figure \ref{relation-example}, the sequential distance from 
\begin{CJK}{UTF8}{gbsn}
``市(City)'' or ``市长(Mayor)'' to ``长(Long)' is $1$ under previous position definition. Actually, ``长(Long)'' is included in ``市长(Mayor)'' and right adjacent to ``市(City)'', but absolute position fails to make a distinction.
\end{CJK}
To address this issue, we propose a relative position relation matrix $L\in \mathbb{N}^{N \times N}$ to present such position information. Similar to~\cite{Xiao:ACL2019}, we enumerate all possible relations between each pair of elements $e_{p:q}$ and $e_{k:l}$ in Figure \ref{relation}.
We give a detailed and vivid example in Figure \ref{relation-example}. For two tokens $t_i$ and $t_j$ refering to $e_{p:q}$ and $e_{k:l}$ respectively, the matrix entry $L_{i,j}$ is the pre-defined relation between them, such as $L_{1,2}=r_1$.

More concretely, in order to make $L$ learnable, we first represent $L$ as the relation position embedding, a 3D tensor $\boldsymbol{\rm{R}}\in \mathbb{R}^{N \times N \times d_r}$ by looking up a trainable embedding matrix $\boldsymbol{\rm{A}}\in \mathbb{R}^{8\times d_r}$, where $d_r$ is the relational embedding dimensionality. Note that here we define eight types of embedding instead of seven relations in Figure \ref{relation}.
The additional embedding is introduced to represent the interaction relation with a shared pivot node (described in the next section) and
facilitate parallel computation.
Then, to incorporate such position relations into attention layer, we adapt Equation \ref{equ:self-att} as follows:

\begin{small}
\begin{gather}
{\rm{\alpha}}\!=\!{\rm{Softmax}}(\frac{\boldsymbol{\rm{Q}}\boldsymbol{\rm{K}}^{T}\!+\!\boldsymbol{\rm{einsum}}(\!``\!ik\!,\!ijk\!\rightarrow\! ij\!"\!,\!\boldsymbol{\rm{Q}}\!,\!\boldsymbol{\rm{R}}^K)}{\sqrt{d_k}})\\
\label{equ:latt}
{\rm {Att}}(\boldsymbol{\rm{Q}}\!,\boldsymbol{\rm{K}}\!,
\!\boldsymbol{\rm{V}})\!=\!{\rm{\alpha}}\boldsymbol{\rm{V}}\!+\!\boldsymbol{\rm{einsum}}(``ik,ikj\!\rightarrow\! ij"\!,\!{\rm{\alpha}},\!\boldsymbol{\rm{R}}^V),
\end{gather}
\end{small}
where $\boldsymbol{\rm{R}}^K\in \mathbb{R}^{N \times N \times d_k}$ and $\boldsymbol{\rm{R}}^V\in \mathbb{R}^{N \times N \times d_v}$ are two relation embedding tensors which are added to the keys and values respectively to indicate relation between input tokens.
In our case, $\boldsymbol{\rm{Q}}$ is a 2D array of shape $[N\times d_k]$ while $\boldsymbol{\rm{R}}^K$ is a 3D array and we need to result in a new array of shape $[N\times N]$, with the element in $i$-th row and $j$-th column is $\sum_{k=1}^{d_k} \boldsymbol{\rm{Q}}_{ik}\boldsymbol{\rm{R}}^K_{ijk}$. To implement this operation, we apply \textbf{einsum}$\footnote{This operator is available in Numpy, TensorFlow, and Pytorch.}$ to sum out the dimension of the hidden size, which is an operation computing multilinear expressions (i.e., sums of products) using the Einstein summation convention.

\paragraph{Porous Multi-Head Attention (PMHA)}
Considering that standard self-attention mechanism encodes sequences by relating sequence items to another one through computation of pairwise similarity, it disperses the distribution of attention and overlooks the local knowledge provided by neighboring elements, which is crucial for NER.
To maintain the strength of capturing long distance dependencies and enhance the ability of capturing short-range dependencies, we sparsify the transformer architecture by replacing the fully-connected topology with a pivot-shared structure referenced by~\cite{Guo:NAACL2019}. Specifically, given element set $E$ and its embedding matrix $\boldsymbol{\rm{X}}$, where $e_{i:j}\in E$ and $\boldsymbol{\rm{x}}_{i:j} \in \boldsymbol{\rm{X}}$ (if $e_{i:j}$ is a character then $\boldsymbol{\rm{x}}_{i:j}=\boldsymbol{\rm{x}}^c_{i}$ else $\boldsymbol{\rm{x}}_{i:j}=\boldsymbol{\rm{x}}^w_{i:j}$ ),
 we define $e_{i:j}^{r_k}$ as the element set whose relation with $e_{i:j}$ is $r_k$,
 $\boldsymbol{\rm{x}}_{i:j}^{r_k}$ as the concatenation of the embeddings where each embedding represents the corresponding element in $e_{i:j}^{r_k}$.
 we also define the neighboring set of $e_{i:j}$ as $\varepsilon\!=\!\{e_{i:j}^{r_1};e_{i:j}^{r_2};e_{i:j}^{r_3};e_{i:j}^{r_4};e_{i:j}^{r_5};e_{i:j}^{r_6}\}$ , then we update the hidden state $\boldsymbol{\rm{h}}_{i:j}$ of $e_{i:j}$ with multi-head attention as follows:
\begin{equation}
\small
\begin{split}
\boldsymbol{\rm{h}}_{i:j}&=[\boldsymbol{\rm{z}}^1_{i:j}; \boldsymbol{\rm{z}}^2_{i:j}; ... ;\boldsymbol{\rm{z}}^H_{i:j}]\boldsymbol{\rm{W}}^O\\
\boldsymbol{\rm{z}}_{i:j}^h&={\rm {Att}}(\boldsymbol{\rm{x}}_{i:j}\boldsymbol{\rm{W}}^Q_h,\boldsymbol{\rm{c}}_{i:j}\boldsymbol{\rm{W}}^K_h,\boldsymbol{\rm{c}}_{i:j}\boldsymbol{\rm{W}}^V_h),\;h\in [1,H]\\
\boldsymbol{\rm{c}}_{i:j}&=[\boldsymbol{\rm{x}}_{i:j}^{r_1}; \boldsymbol{\rm{x}}_{i:j}^{r_2}; \boldsymbol{\rm{x}}_{i:j}^{r_3};\boldsymbol{\rm{x}}_{i:j}^{r_4}; \boldsymbol{\rm{x}}_{i:j}^{r_5}; \boldsymbol{\rm{x}}_{i:j}^{r_6}; \boldsymbol{\rm{s}}]\\
\boldsymbol{\rm{s}}&=\frac{1}{n}\sum_{i,j}\boldsymbol{\rm{x}}_{i:j},
\end{split}
\end{equation}
where $\boldsymbol{\rm{W}}^Q_h, \boldsymbol{\rm{W}}^K_h, \boldsymbol{\rm{W}}^V_h$ are trainable projection matrices corresponding to the $h$-th head, $\boldsymbol{\rm{z}}_h$ is the $h$-th output, $H$ is the number of heads and Att() is defined in Equation \ref{equ:latt}. 
As we can see, in our porous multi-head attention,
one element $e_{i:j}$ just makes direct attention computation with its neighboring elements and models the non-local compositions via the pivot node $\boldsymbol{\rm{s}}$.
As illustrated in Figure \ref{relation-example}, $e_{i:j}$ doesn't compute attention directly with the element set $e_{i:j}^{r_7}$, thus we mask them.
Under this lightweight porous structure, our transformer encoder has an approximate ability to strengthen local dependencies among neighboring tokens and keep the ability to capture long distance dependencies.

\subsection{BiGRU-CRF Decoding}

After extracting the semantic information by the porous lattice transformer encoder layer, we feed the character sequence representations into a BiGRU-CRF decoding layer to make sequence tagging. Specifically, taking $[\boldsymbol{\rm{x}}_1^c;\boldsymbol{\rm{h}}_{1:1}],...,[\boldsymbol{\rm{x}}_n^c;\boldsymbol{\rm{h}}_{n:n}]$ as input, a bidirectional GRU is implemented to produce forward state $\overrightarrow{\boldsymbol{\rm{h}}}_t$ and backward state $\overleftarrow{\boldsymbol{\rm{h}}}_t$ for each time step, and then we concatenate these two separate hidden states as the encoding output of the $t$-th character, donated as $\boldsymbol{\rm{h}}_t=[\overrightarrow{\boldsymbol{\rm{h}}}_t;\overleftarrow{\boldsymbol{\rm{h}}}_t]$.

Finally, a standard CRF layer is used on top of $\boldsymbol{\rm{h}}_1,\boldsymbol{\rm{h}}_2,...,\boldsymbol{\rm{h}}_n$ to make sequence tagging. For a label sequence $\boldsymbol{\rm{y}}=\{y_1,y_2,...,y_n\}$, we define its probability to be:
\begin{equation}
\small
P(\boldsymbol{\rm{y}}|S)\!=\!\frac{{\rm{exp}}(\sum_i(\boldsymbol{\rm{W}}_{\text{CRF}}^{y_i}\boldsymbol{\rm{h}}_i\!+\!b_{\text{CRF}}^{(y_{i-1},y_i)}))}{\sum_{y'}{\rm{exp}}(\sum_i(\boldsymbol{\rm{W}}_{\text{CRF}}^{y_i'}\boldsymbol{\rm{h}}_i\!+\!b_{\text{CRF}}^{(y_{i-1}',y_i')}))},
\end{equation}

where $y'$ denotes all possible tag sequences, $\boldsymbol{\rm{W}}_{\text{CRF}}^{y_i}$ is a model parameter specific to $y_i$, and $b_{\text{CRF}}^{(y_{i-1},y_i)}$ is the transition score between $y_{i-1}$ and $y_i$. For decoding, we use the first-order Viterbi algorithm to find the label sequence that obtains highest score.

\subsection{Training}
Given a set of manually labeled training data $\{(S_i,\boldsymbol{\rm{y}}_i)\}|_{i=1}^N$, sentence-level log-likelihood loss with $L_2$ regularization is used to train the model:
\begin{small}
\begin{equation}
\mathcal{L}=\sum_{i=1}^N {\rm{log}}(P(\boldsymbol{\rm{y}}_i|S_i))+\frac{\lambda}{2}\parallel \Theta\parallel^2,
\end{equation}
\end{small}
where $\lambda$ is the $L_2$ regularization weight and $\Theta$ represents the parameter set.

\section{Experiments}
We conduct experiments to investigate the effectiveness of our proposed PLTE method across different domains. Standard precision (P), recall (R) and F1-score (F1) are used as evaluation metrics.
\subsection{Experimental Setup}
\subsubsection{Data}
We evaluate our model on four datasets, including OntoNotes~\cite{Ralph:2011}, MSRA~\cite{Gina-Anne:SIGHAN2006}, Weibo NER~\cite{Peng:EMNLP2015,He:EACL2017} and a Chinese Resume dataset~\cite{Zhang:ACL2018}. We use the same training, valid and test split as~\cite{Zhang:ACL2018}.
For these datasets, both OntoNotes and MSRA are in news domain, while  Weibo and Resume come from social media.

\subsubsection{Baseline Methods}
We compare our proposed model to several recent lexicon-enhanced character-based models. 

\textbf{Lattice LSTM.} Lattice LSTM~\cite{Zhang:ACL2018} exploits lexical information in character sequence through gated recurrent cells, which can avoid segmentation errors.

\textbf{LR-CNN.} LR-CNN~\cite{Gui:2019IJCAI} is the latest SOTA method of Chinese NER, which incorporates lexicons using a rethinking mechanism.

Furthermore, to explore the effectiveness of pre-trained language model, we implement several baselines based on BERT representations.

\textbf{BERT-Tagger.} BERT-Tagger~\cite{devlin:2019naacl} uses the outputs from the last layer of model $\text{BERT}_{\emph{base}}$ as the character-level enriched contextual representations to make sequence labelling.

\textbf{PLTE[BERT]/LR-CNN[BERT]/Lattice LSTM[BERT].} These three models replace character embeddings with the pre-trained BERT representations, and use softmax layer to make sequence tagging.

\subsubsection{Hyper-parameter settings}
In our experiments, we use the same character embeddings, character bigram embeddings and word embeddings as~\cite{Zhang:ACL2018}, which are pre-trained on Chinese Giga-word \footnote{\url{https://catalog.ldc.upenn.edu/ LDC2011T13}} using Word2vec~\cite{Mikolov:NIPS2013} and fine-tuned during training.
The model is trained using stochastic gradient descent with the initial learning rate of 0.045 and the weight decay of 0.05.
Dropout is applied to the embeddings and GRU layer with a rate of 0.5 and the transformer encoder with 0.3.
For the biggest dataset MSRA and the smallest dataset Weibo, we set the dimensionality of GRU hidden states as 200 and 80 respectively.
For the other datasets, this dimension is set to 100.
What's more, the hidden size and the number of heads are set to 128 and 6, respectively.
For models based on BERT, we fine-tune BERT representation layer during training. 
We use BertAdam to optimize all trainable parameters, select the best learning rate from 1$e$-5 to 1$e$-4 on the development set.

\begin{table}[t]
\begin{floatrow}
\capbtabbox{
\small
  \begin{tabular}{|l|llll|}

        \hline
        \multicolumn{5}{|c|}{OntoNotes}\\
        \hline
        \textbf{Input} & \textbf{Models} & \textbf{P} & \textbf{R} & \textbf{F1} \\
        \hline
        \multirow{4}{*}{Gold seg} & \newcite{Che:2013NAACL}     & 77.71 & 72.51 & 75.02 \\
                                  & \newcite{Wang:2013AAAI}    & 76.43 & 72.32 & 74.32 \\
                                  & \newcite{Yang:2016CICLing} & 72.98 & 80.15 & \textbf{76.40} \\
        \hline
        \multirow{8}{*}{No seg}   
                                  & Lattice LSTM \shortcite{Zhang:ACL2018}                             & 76.35 & 71.56 & 73.88 \\
                                  & LR-CNN \shortcite{Gui:2019IJCAI}                                   & 76.40 & 72.60 & 74.45 \\
                                  &CAN-NER\shortcite{Zhu:2019NAACL}                 &75.05 & 72.29 & 73.64 \\
                                  &PLTE                                 & 76.78 & 72.54 & \textbf{74.60} \\
        \cline{2-5}
                                  &BERT-Tagger                          & 78.01 & 80.35 & 79.16 \\
                                  &Lattice LSTM[BERT]                   & 79.79 & 79.41 & 79.60 \\
                                  &LR-CNN[BERT]                         & 79.41 & 80.32 & 79.86 \\
                                  &PLTE[BERT]                           & 79.62 & 81.82 & \textbf{80.60} \\
         \hline
         \hline

        \multicolumn{5}{|c|}{Resume}\\
        \hline
        \multicolumn{2}{|l}{\textbf{Models}} & \textbf{P} & \textbf{R} & \textbf{F1} \\
        \hline
        \multicolumn{2}{|l}{Lattice LSTM \shortcite{Zhang:ACL2018}} & 94.81 & 94.11 & 94.46 \\
        \multicolumn{2}{|l}{CAN-NER\shortcite{Zhu:2019NAACL}}       & 95.05 & 94.82 & 94.94 \\
        \multicolumn{2}{|l}{LR-CNN \shortcite{Gui:2019IJCAI}}       & 95.37 & 94.84 & 95.11 \\

        \multicolumn{2}{|l}{PLTE}                                   & 95.34 & 95.46 & \textbf{95.40} \\
        \hline
        \multicolumn{2}{|l}{BERT-Tagger}                            & 96.12 & 95.45 & 95.78 \\
        \multicolumn{2}{|l}{Lattice LSTM[BERT]}                     & 95.79 & 95.03 & 95.41 \\
        \multicolumn{2}{|l}{LR-CNN[BERT]}                           & 95.68 & 96.44 & 96.06 \\
        \multicolumn{2}{|l}{PLTE[BERT]}                             & 96.16 & 96.75 & \textbf{96.45} \\
        \hline
    \end{tabular}
    }{
    \caption{Main results on OntoNotes and Resume}
 \label{Ontonotes}}

\capbtabbox{
\small
  \begin{tabular}{|llll|}

        \hline
        \multicolumn{4}{|c|}{MSRA}\\
        \hline
        \textbf{Models} & \textbf{P} & \textbf{R} & \textbf{F1} \\
        \hline
        \newcite{Zhou:2013Chinese}  & 91.86 & 88.75 & 90.28 \\
        \newcite{Lu:LREC2016}        & - & - & 87.94 \\
        \newcite{Cao:2018AAAI}& 91.73 & 89.58 & 90.64 \\
        Lattice LSTM \shortcite{Zhang:ACL2018}                                    & 93.57 & 92.79 & 93.18 \\
        CAN-NER\shortcite{Zhu:2019NAACL}& 93.53 & 92.42 & 92.97 \\
        LR-CNN \shortcite{Gui:2019IJCAI}                                         & 94.50 & 92.93 & \textbf{93.71} \\
        PLTE                              & 94.25 & 92.30 & 93.26 \\
        \hline
        BERT-Tagger                       & 94.43 & 93.86 & 94.14 \\
        Lattice LSTM[BERT]                & 93.99 & 92.86 & 93.42 \\
        LR-CNN[BERT]                      & 94.68 & 94.03 & 94.35 \\
        PLTE[BERT]                        & 94.91 & 94.15 & \textbf{94.53} \\
        \hline
        \hline
        \multicolumn{4}{|c|}{Weibo}\\
        \hline
        \textbf{Models} & \textbf{P} & \textbf{R} & \textbf{F1} \\
        \hline
        \newcite{Peng:2016ACL}    & 66.47 & 47.22 & 55.28\\
        \newcite{He:2017AAAI}        & 61.68 & 48.82 & 54.50 \\
        \newcite{Cao:2018AAAI}       & 59.51 & 50.00 & 54.43 \\
        Lattice LSTM \shortcite{Zhang:ACL2018} &  52.71 & 53.92 & 53.13 \\
        LR-CNN \shortcite{Gui:2019IJCAI}       &  65.06 & 50.00 & \textbf{56.54} \\

        PLTE                                   & 62.21 & 49.54 & 55.15 \\
        \hline
        BERT-Tagger                            & 67.12 & 66.88 & 67.33 \\
        Lattice LSTM[BERT]                     & 61.08 & 47.22 & 53.26\\
        LR-CNN[BERT]                           & 64.11 & 67.77 & 65.89 \\
        PLTE[BERT]                             & 72.00 & 66.67 & \textbf{69.23} \\
        \hline
    \end{tabular}
    }{
    \caption{Main results on MSRA and Weibo}
 \label{msra}}

    \end{floatrow}

\end{table}

\subsection{Results}
\textbf{OntoNotes.} Table \ref{Ontonotes} illustrates our experimental results on OntoNotes. The ``Input" column indicates whether the input sentences are segmented or not, where methods in \textbf{Gold seg} process word sequences with gold segmentation and \textbf{No seg} indicates that the input sentence is a character sequence. 

With gold-standard segmentation, all of the word-level models~\cite{Che:2013NAACL,Wang:2013AAAI,Yang:2016CICLing} achieve strong performance by using segmentation and external labeled data. But such information is not available in most datasets, such that we only use pre-trained character and word embeddings as our resource.

Under No-segmentation settings, we first compare 3 widely-used non-BERT models. Our PLTE model achieves the best F1 score and gains a  $0.72\%$ improvement over lattice LSTM in F1 score since our model integrates lexical words information into self-attention computation in a more effective way.\footnote{Case study is provided in Table 1 in the Appendix.} With pre-trained $\text{BERT}_{\emph{base}}$, BERT-Tagger leads to a significant boost in performance to $79.16\%$. On this basis, our proposed PLTE[BERT] model outperforms the BERT-Tagger by $1.44\%$ in F1 score on OntoNotes.

\textbf{MSRA/Weibo/Resume}
Tables \ref{Ontonotes} and \ref{msra} present comparisons among various methods on the MSRA, Weibo, and Resume datasets.
Existing statistical methods explore the rich statistical features~\cite{Zhou:2013Chinese} and character embedding features~\cite{Lu:LREC2016}.
For neural models, some existing models use multi-task learning~\cite{Peng:2016ACL,Cao:2018AAAI} or semi-supervised learning~\cite{He:2017AAAI}. 
CAN-NER~\cite{Zhu:2019NAACL} investigate a character-based convolutional attention network coupled with GRU for Chinese NER.

\begin{table*}
 \centering
 \small
 \begin{tabular}{c|c c c||c c c}
  \hline
  \hline
  \multirow{2}{*}{\textbf{Models}} & \multicolumn{3}{c||}{\textbf{Word2vec}} & \multicolumn{3}{c}{\textbf{BERT}}  \\
  \cline{2-7}
  ~& Lattice LSTM  & LR-CNN  &  PLTE & Lattice LSTM  & LR-CNN  &  PLTE \\
  \hline
  OntoNotes   & $1\times$ & $2.23\times$      & $11.4\times$ & $1\times$      & $1.96\times$ & $6.21\times$           \\
  \hline
  MSRA        & $1\times$ & $1.57\times$      & $8.48\times$ & $1\times$      & $1.97\times$ & $7.11\times$        \\
  \hline
  Weibo       & $1\times$ & $2.41\times$      & $9.12\times$ & $1\times$      & $2.02\times$ & $6.48\times$        \\
  \hline
  Resume      & $1\times$ & $1.44\times$      & $9.68\times$ & $1\times$      & $1.46\times$ & $5.57\times$        \\
  \hline
 \end{tabular}
 \caption{\label{efficiency-table}
Testing-time speedup of different models. Lattice LSTM and LR-CNN can only run with $batch\_size\!=\!1$ while our PLTE model runs with $batch\_size\!=\!16$.
}
\end{table*}

Consistent with observations on OntoNotes, 
all the lexicon-enhanced methods achieve higher F1 scores than character-based methods, which demonstrates the usefulness of lexical word information.
With pre-trained contextual representations, BERT-based models outperform non-BERT models by a large margin.
Even though the original BERT model already provides strong prediction power, PLTE consistently improves over BERT-Tagger, lattice LSTM[BERT] and LR-CNN[BERT], which indicates that our proposed PLTE model can make better use of these semantic representations.
Another interesting observation is that PLTE gains more significant improvement when combined with BERT compared with other lexicon-enhanced methods.
We suspect that it is because PLTE is more capable of fully leveraging the language information embedded in the input representations.
While the embeddings pre-trained by Word2vec are not as informative to PLTE to fulfill its  potential, BERT representation can well capture rich semantic patterns and help PLTE improve the performance.
\begin{figure*}[t]
\subfigure[]{
\label{len-compare}
\includegraphics[width=0.46\textwidth]{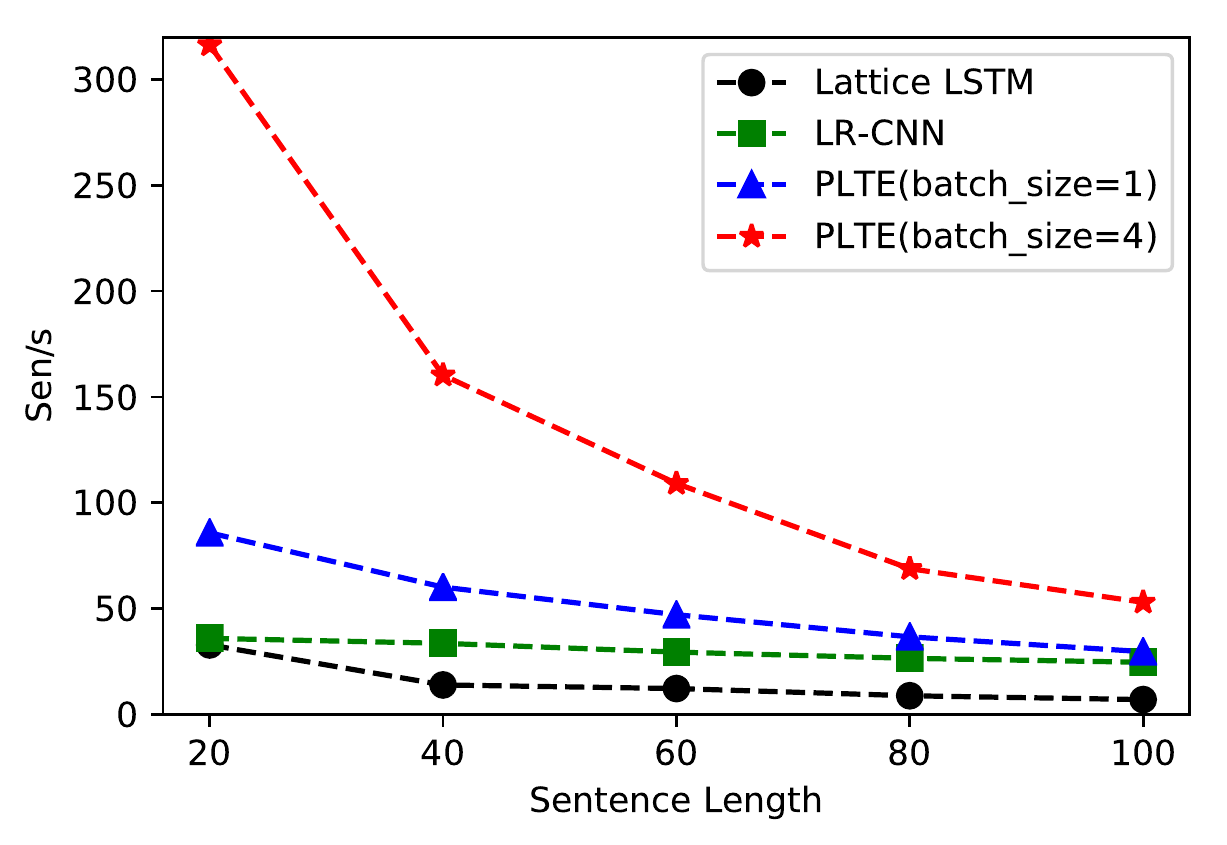}
}
\subfigure[]{
\label{ablationStudy}

\includegraphics[width=0.46\textwidth]{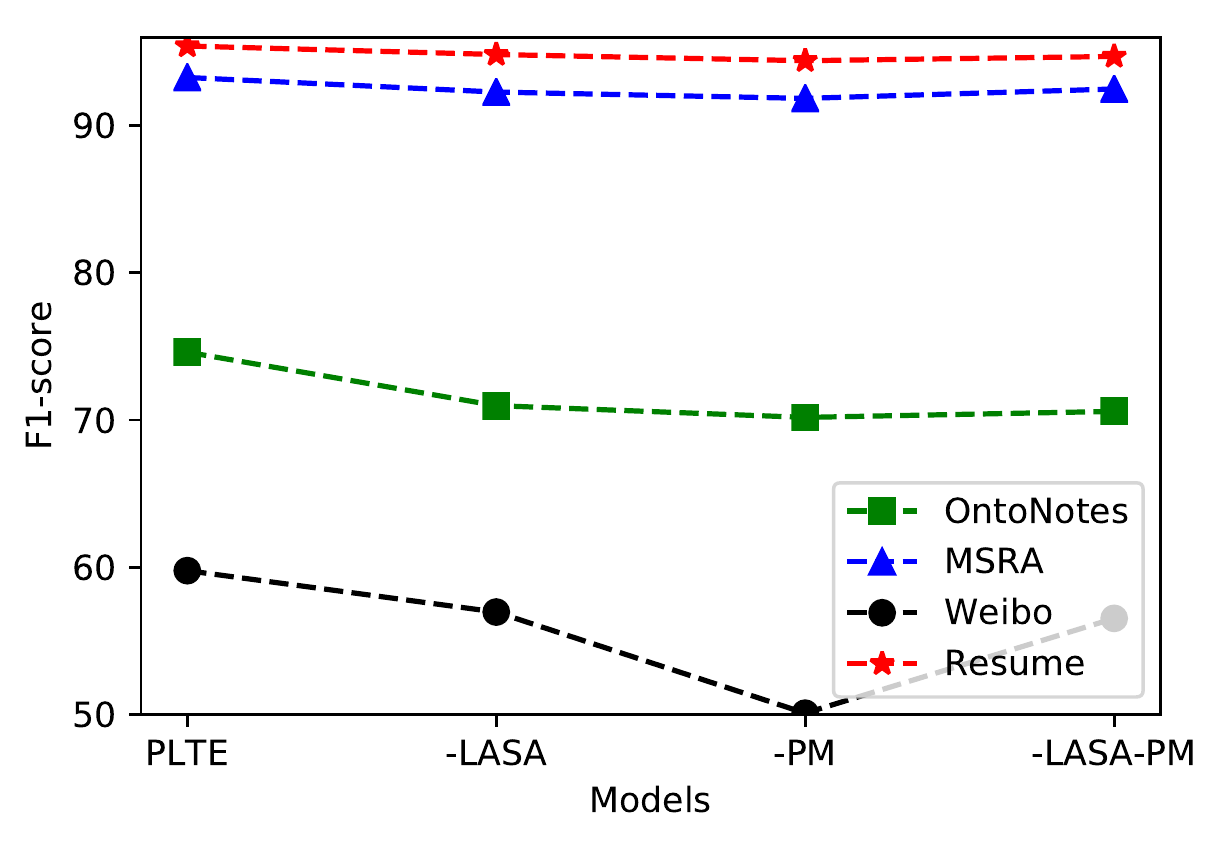}
}%
\centering
\caption{
(a) Test speed against the sentence length. Sen/s denotes the number of sentences processed per second;
and (b) An ablation study of our proposed model. 
For model without lattice-aware self-attention (-LASA), we take character sequence as input, and one character just computes multi-head self-attention weights with its adjacent characters and the shared pivot node.
For model without porous mechanism (-PM), we directly utilize multi-head LASA to aggregate the weighted information of each word with fully-connected attention connections.
For PLTE-LASA-PMHA, we apply multi-head self-attention to each pair of elements from the input character sequence.
}

\end{figure*}

\subsection{Experimental Discussion}
\subsubsection{Efficiency Advantage}
PLTE also outperforms current lexicon-enhanced methods in efficiency. Table \ref{efficiency-table} lists the test times of different models with different input representations on all four benchmarks.
As we can see, PLTE runs up to 11.4 and 5.11 times faster than lattice LSTM and LR-CNN respectively with Word2vec embeddings on OneNotes.
Similar efficiency improvement can also be observed on other datasets under both two kinds of input representations.
Aligning word-character lattice structure for batch running can be usually non-trivial~\cite{sui:2019emnlp} and both lattice LSTM and LR-CNN have no ability in batch-running due to the DAG structure or variable-sized lexical words set. In contrast, PLTE overcomes this limitation since we can simply concatenate all the elements as input thanks to the lattice-aware self-attention mechanism, which calculates attention weights between each pair of tokens by matrix multiplication, thus can be computed parallelly in batches.

To investigate the influence of the different sentence lengths, we conduct experiments on OntoNotes by splitting this dataset into five parts according to sentence length.
The results in Figure \ref{len-compare} demonstrate that PLTE runs faster than lattice LSTM and LR-CNN with different sentence lengths, especially for short sentences.
In particular, when the sentence length is less than 20, PLTE($batch\_size\!=\!4$) runs 9.64 times faster than lattice LSTM and 8.81 times faster than LR-CNN.
When the sentence length increases, the efficiency gains from batching computation decline gradually due to the limited computing resources of a single GPU.
Besides, even if we set the batch size as 1, PLTE still has remarkable advantage in speed, since lattice LSTM demands multiple recurrent computation steps, and the rethinking mechanism in LR-CNN is also computationally expensive. 

\subsubsection{Model Ablation study}
We conduct an ablation study on four datasets to understand the effectiveness of each component, the results are shown in Figure~\ref{ablationStudy}.
We can observe that:
(1) Removing the LASA module hurts the results by $3.63\%$, $0.99\%$, $2.81\%$ and $0.58\%$ F1 score on four datasets respectively, which indicates that lexicons play an important role in character-level Chinese NER.
(2) By introducing the porous mechanism (PM), we can enhance the ability of capturing useful local context, which is beneficial to NER, while maintaining the strength of capturing long-term dependencies.
(3) PLTE-PM performs worse than PLTE-LASA-PM, which confirms that the standard LASA is not suitable for NER because it takes into account all the signals and disperses the distribution of attention, while NER may be benefited more from local modeling. 
(4) PLTE-LASA outperforms PLTE-LASA-PM on most datasets, which shows that the porous mechanism can also benefit self-attention when only taking characters as input.

\section{Conclusion}
We presented PLTE, a porous lattice transformer encoder which incorporates lexicons into character-level Chinese NER.
PLTE enables the interaction between the matched lexical words and their constituent characters, and proceeds in batches with the lattice-aware self-attention.
It also learns a porous attention distribution to enhance the ability of localness modeling.
We evaluate the proposed model on four Chinese NER datasets. 
Using Word2vec embeddings, our PLTE outperforms various baselines and performs up to 11.4 times faster than previous lattice-based method.
Switching to BERT representations, PLTE achieves more significant performance gain than existing methods.
There are multiple venues for future work, where one promising direction is to apply our model to the pre-training procedure of Chinese Transformer language models.

\bibliographystyle{coling}
\bibliography{coling2020}

\end{document}